\documentclass[letterpaper, 10 pt, journal, twoside]{IEEEtran}
\usepackage{amsmath}
\usepackage{cite}
\usepackage{color}
\usepackage{epsfig}
\usepackage{epstopdf}
\usepackage[T1]{fontenc}
\usepackage{graphicx}
\usepackage{subfigure} 
\usepackage{mathrsfs}
\usepackage{makecell}
\usepackage{multirow}
\usepackage{times}
\usepackage{epsfig}
\usepackage{graphicx}
\usepackage{amsmath}
\usepackage{amssymb}
\usepackage{cuted}
\usepackage{capt-of}
\usepackage{subfigure}
\usepackage{multirow}
\usepackage{makecell}
\usepackage{tabulary}
\usepackage{booktabs}
\usepackage{array}
\usepackage{url}
\usepackage[table]{xcolor}
\usepackage{float}
\usepackage{makecell}
\usepackage{balance}

\DeclareGraphicsExtensions{.pdf,.png,.jpg,.fig}
\usepackage[linesnumbered,ruled]{algorithm2e}

\newcommand{\ie}{\textit{i.e.}}
\newcommand{\eg}{\textit{e.g.}}
\newcommand{\etc}{\textit{etc}}
\newcommand{\etal}{\textit{et al.}}

\usepackage[pagebackref=false,breaklinks=false,colorlinks,bookmarks=false,citecolor=black]{hyperref}

\begin{document}

\title{Three-Filters-to-Normal: An Accurate and Ultrafast Surface Normal Estimator}
\author{
	Rui~Fan,~\IEEEmembership{Member,~IEEE},~Hengli~Wang,~\IEEEmembership{Graduate Student Member,~IEEE},\\Bohuan~Xue,~\IEEEmembership{Graduate Student Member,~IEEE},~Huaiyang~Huang,~\IEEEmembership{Graduate Student Member,~IEEE},\\Yuan~Wang,~Ming~Liu,~\IEEEmembership{Senior Member,~IEEE},~Ioannis~Pitas,~\IEEEmembership{Fellow,~IEEE}
	\thanks{
		\vspace{-1em}
		
		R. Fan is with the Department of Computer Science and Engineering as well as the Department of Ophthalmology, the University of California San Diego, La  Jolla, CA 92093, United States. (e-mail: rui.fan@ieee.org)
		
		H. Wang, B. Xue, H. Huang and M. liu are with the Robotics Institute, the Hong Kong University of Science and Technology, Hong Kong SAR, China. (e-mail: \{hwangdf, bxueaa, hhuangat, eelium\}@ust.hk)
		
		Y. Wang is with the Industrial R\&D Center, SmartMore, Shenzhen 518040, China.
		(e-mail: yuan.wang@smartmore.com)
		
		I. Pitas is with the School of Informatics, Aristotle University of Thessaloniki, Thessaloniki 541 24, Greece (e-mail: pitas@aiia.csd.auth.gr).
		
		R. Fan, H. Wang and B. Xue contributed equally to this work.
	}
	\vspace{-2.0em}
}
\markboth{IEEE Robotics and Automation Letters}
{Fan \MakeLowercase{\textit{et al.}}: Three-Filters-to-Normal: An Accurate and Ultrafast Surface Normal Estimator}
\maketitle

\begin{strip}
\vspace{-8.0em}
\centering
\includegraphics[width=0.99999\textwidth]{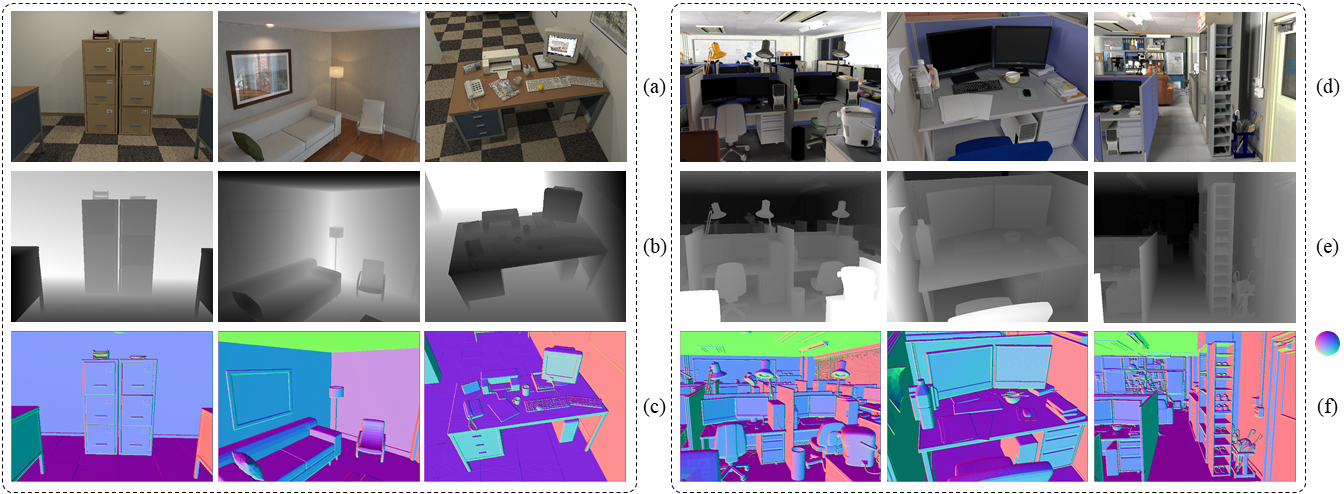}
\captionof{figure}{
		Surface normal estimation from depth and disparity images: (a) and (b) show examples of RGB and depth images of the Augmented ICL-NUIM dataset \cite{Choi_2015_CVPR}, respectively; (d) and (e) show examples of RGB and disparity images of the Tsukuba stereo dataset \cite{martull2012realistic}, respectively; (c) and (f) show the surface normals estimated from (b) and (e), respectively, using 3F2N SNE.
}
\label{fig.demo_fig}
\end{strip}

\begin{abstract}
This paper proposes three-filters-to-normal (3F2N), an accurate and ultrafast surface normal estimator (SNE), which is designed for structured range sensor data, \eg,  depth/disparity images. 3F2N SNE computes surface normals by simply performing three filtering operations (two image gradient filters in horizontal and vertical directions, respectively, and a mean/median filter) on an inverse depth image or a disparity image. Despite the simplicity of 3F2N SNE, no similar method already exists in the literature. To evaluate the performance of our proposed SNE, we created three large-scale synthetic datasets (easy, medium and hard) using 24 3D mesh models, each of which is used to generate 1800--2500 pairs of depth images (resolution: 480$\times$640 pixels) and the corresponding ground-truth surface normal maps from different views. 3F2N SNE demonstrates the state-of-the-art performance, outperforming all other existing geometry-based SNEs, where the average angular errors with respect to the easy, medium and hard datasets are 1.66$\boldsymbol{^\circ}$, 5.69$\boldsymbol{^\circ}$ and 15.31$\boldsymbol{^\circ}$, respectively. Furthermore, our C++ and CUDA implementations achieve a processing speed of over 260 Hz and 21 kHz, respectively. Our datasets and source code are publicly available at \url{sites.google.com/view/3f2n}.
\end{abstract}

\section{Introduction}
\label{sec.introduction}
Real-time 3-dimensional (3D) object recognition is a very challenging computer vision task \cite{hinterstoisser2011gradient}. Surface normal is an informative and important visual feature used in 3D object recognition \cite{klasing2009comparison}. However, not much research has been conducted thoroughly on surface normal estimation, as it is merely considered as an auxiliary functionality for other computer vision applications. Such applications are generally required to perform in an online fashion, and therefore, surface normal estimation must be carried out extremely fast \cite{klasing2009comparison}.

The surface normals can be estimated from either a 3D point cloud or a depth/disparity image (see Fig. \ref{fig.demo_fig}). The former, such as a LiDAR point cloud, is generally unstructured. Estimating surface normals from unstructured range data usually requires the generation of an undirected graph \cite{klasing2009comparison}, \eg, a $k$-nearest neighbor graph or a Delaunay tessellation graph. However, the generation of such graphs is very computationally intensive. Therefore, in recent years, many researchers have been focusing on surface normal estimation from structured range sensor data, \eg, depth/disparity images.

Existing surface normal estimators (SNEs) can be categorized as either geometry-based \cite{hinterstoisser2011gradient, klasing2009comparison, badino2011fast, lu2017symps} or machine/deep learning-based \cite{bansal2016marr, li2015depth}.  The former typically computes surface normals by fitting planar or curved surfaces to locally selected 3D point sets, using statistical analysis or optimization techniques, \eg, singular value decomposition (SVD) or principal component analysis (PCA) \cite{klasing2009comparison}. On the other hand, the latter generally utilizes data-driven classification/regression models, \eg, convolutional neural networks (CNNs) to infer surface normal information from RGB or depth images \cite{qi2018geonet}.

In recent years, with rapid advances in machine/deep learning, many researchers have resorted to deep convolutional neural networks (DCNNs) for surface normal estimation. For instance, Xu \etal \cite{xu2018pad} utilized a so-called prediction-and-distillation network (PAD-Net) to 1) realize monocular depth prediction and surface normal inference, as well as 2) perform scene parsing and contour detection simultaneously.
Recently, Huang \etal \cite{huang2019framenet} formulated the problem of densely estimating local 3D canonical frames from a single RGB image as a joint estimation of surface normals, canonical tangent directions and projected tangent directions. Such problem was then addressed by a DCNN.

The existing data-driven SNEs are generally trained using supervised learning techniques. Hence, they require a large amount of hand-labeled training data to find the best CNN parameters \cite{li2015depth}. Additionally, such CNNs were not specifically designed for surface normal estimation, because SNEs were only used as an auxiliary functionality for other computer vision applications, such as scene parsing, 3D object detection, and depth perception. Furthermore, many robotics and computer vision applications, \eg, autonomous driving \cite{fan2020sne}, require very fast surface normal estimation (in milliseconds). Unfortunately, the existing machine/deep learning-based SNEs are not fast enough. Moreover, the  accuracy achieved by data-driven SNEs is still far from satisfactory (the average proportion of good pixels is usually lower than $80\%$) \cite{bansal2016marr, li2015depth}. Most importantly, it can be considered more reasonable to estimate surface normals from point clouds or disparity/depth images rather than from RGB images. Hence, there is a strong motivation to develop a lightweight SNE for structured range data with high accuracy and speed.

The major contributions of this work are as follows:

\begin{enumerate}
\item \textbf{Three-filter-to-normal (3F2N)}, an accurate and ultrafast SNE. We published its Matlab, C++ and CUDA implementations at \url{github.com/ruirangerfan/three_filters_to_normal}. Compared with other geometry-based SNEs, 3F2N SNE greatly improves the trade-off between speed and accuracy. 
\item Three datasets (easy, medium and hard) created using 24 3D mesh models. Each mesh model is used to generate 1800--2500 depth images from different views. The corresponding surface normal ground truth is also provided, as 3D mesh object models (rather than the objects themselves) are available for surface normal ground truth generation.
\end{enumerate}

\section{Related Work}
This section provides an overview of geometry-based SNEs.

\label{sec.related_work}
1) {PlaneSVD} SNE \cite{jordan2014quantitative}: The simplest way to estimate the surface normal of an observed 3D point  $\mathbf{p}_i=[x,y,z]^\top$ in the camera coordinate system (CCS) is to fit a local plane:
\begin{equation}
	n_x x+n_y y+n_z z+b=0
	\label{eq.plane_function}
\end{equation}
to the points in $\mathbf{Q}_i^+=[\mathbf{Q}_i^\top, \mathbf{p}_i]^\top$, where $\mathbf{Q}_i=[{\mathbf{q}_i}_1,\dots, {\mathbf{q}_i}_k]^\top$ (${\mathbf{q}_i}_j\neq\mathbf{p}_i$) is a set of $k$ neighboring points of $\mathbf{p}_i$. The surface normal $\mathbf{n}_i=[n_x,n_y,n_z]^\top$ can be estimated by solving:
\begin{equation}
	\min_{\mathbf{b}_i} \Big|\Big| \Big[\mathbf{Q}_i^+ \ \ \mathbf{1}_{k+1} \Big] \mathbf{b}_i \Big|\Big|_2,
	\label{eq.planesvd}
\end{equation}
where $\mathbf{b}_i=[\mathbf{n}_i^\top, b]^\top$ and $\mathbf{1}_m$ is an $m$-entry vector of ones. (\ref{eq.plane_function}) can be solved by factorizing $\mathbf{Q}_i^+$ into $\mathbf{U}\mathbf{\Sigma}\mathbf{V}^\top$ using SVD. $\hat{\mathbf{b}}_i$ (the optimum $\mathbf{b}_i$) is a column vector in $\mathbf{V}$ corresponding to the smallest singular value in $\mathbf{\Sigma}$ \cite{klasing2009comparison}.

2) {PlanePCA} SNE \cite{klasing2009realtime}: $\mathbf{n}_i$ can also be estimated by removing the empirical mean $\bar{\mathbf{q}}_i=\frac{1}{k+1}(\mathbf{p}_i+\Sigma_{j=1}^{k}{\mathbf{q}_{i}}_j)$ from $\mathbf{Q}^+_i$ and rearranging (\ref{eq.planesvd}) as follows \cite{fan2019pothole}:
\begin{equation}
	\min_{\mathbf{n}_i} \Big|\Big| \Big[\mathbf{Q}_i^+ - \bar{\mathbf{Q}}_i^+  \Big] \mathbf{n}_i \Big|\Big|_2,
	\label{eq.planepca}
\end{equation}
where $\bar{\mathbf{Q}}_i^+=\mathbf{1}_{k+1}\bar{\mathbf{q}}_i^\top$. Minimizing (\ref{eq.planepca}) is equivalent to performing PCA on $\mathbf{Q}^+_i$ and selecting the principal component with the smallest covariance \cite{klasing2009comparison}.

3) VectorSVD SNE \cite{klasing2009comparison}: A straightforward alternative to fitting (\ref{eq.plane_function}) to $\mathbf{Q}^+_i$ is to minimize the sum of the inner dot products between ${\mathbf{r}_i}_j={\mathbf{q}_i}_j-\mathbf{p}_i$ and $\mathbf{n}_i$, namely,
\begin{equation}
	\min_{\mathbf{n}_i} \Big|\Big| \Big[\mathbf{Q}_i - \mathbf{1}_k\mathbf{p}_i^\top  \Big] \mathbf{n}_i \Big|\Big|_2.
	\label{eq.vectorSVD}
\end{equation}
This minimization is done by SVD.

4) AreaWeighted SNE \cite{klasing2009comparison}:
A triangle can be formed by a given pair of ${\mathbf{r}_{i}}_j$ and ${\mathbf{r}_{i}}_{j+1}$, as defined above. A general expression of averaging-based SNEs is as follows \cite{klasing2009comparison}:
\begin{equation}
	\mathbf{n}_i=\frac{1}{k}\sum_{j=1}^k w_j \frac{{\mathbf{r}_{i}}_j\times{\mathbf{r}_{i}}_{j+1}}{\|{\mathbf{r}_{i}}_{j}\times{\mathbf{r}_{i}}_{j+1}\|_2},
	\label{eq.averaging_based}
\end{equation}
where $w_j$ is a weight and ${\mathbf{r}_{i}}_{k+1}={\mathbf{r}_{i}}_{1}$. In AreaWeighted SNE, the surface normal of each triangle is weighted by the magnitude of its area:
\begin{equation}
	w_j=\frac{1}{2}\|{\mathbf{r}_{i}}_{j}\times{\mathbf{r}_{i}}_{j+1}\|_2.
	\label{eq.areaweighted}
\end{equation}

5) AngleWeighted SNE \cite{klasing2009comparison}: The weight $w_j$ of each triangle relates to the angle between ${\mathbf{r}_{i}}_{j}$ and ${\mathbf{r}_{i}}_{j+1}$:
\begin{equation}
	w_j=\cos^{-1}\Bigg(\frac{\langle{\mathbf{r}_{i}}_{j}, {\mathbf{r}_{i}}_{j+1}\rangle}{\|{\mathbf{r}_{i}}_{j}\|_2\|{\mathbf{r}_{i}}_{j+1}\|_2}\Bigg),
\end{equation}
where $\langle \cdot \rangle$ is a dot product operator.

6) FALS SNE \cite{badino2011fast}: The relationship between the Cartesian coordinate system and the spherical coordinate system (SCS) is as follows \cite{badino2011fast}:
\begin{equation}
	\mathbf{p}_i=r_i\mathbf{v}_i=r_i\begin{bmatrix}
		\sin\theta_i\cos\phi_i \\
		\sin\phi_i             \\
		\cos\theta_i\cos\phi_i
	\end{bmatrix},
	\label{eq.cartesian_spherical}
\end{equation}
where $r_i\geq0$,  $\theta_i\in(-\pi,\pi]$ and $\phi_i\in(-\frac{\pi}{2},\frac{\pi}{2}]$. Since all points in $\mathbf{Q}_i^+$ are in a small neighborhood \cite{badino2011fast}, their $r_i$ are considered to be identical in FALS SNE. (\ref{eq.planesvd}) and (\ref{eq.cartesian_spherical}) result in:
\begin{equation}
	\min_{\tilde{\mathbf{n}}_i} \Big|\Big| \mathbf{V}_i^+ \tilde{\mathbf{n}}_i - \mathbf{s}_i \Big|\Big|_2,
	\label{eq.fals}
\end{equation}
where $\mathbf{V}_i^+=[\mathbf{v}_i, {\mathbf{v}_{i}}_1,\dots,{\mathbf{v}_i}_k]^\top$, $\tilde{\mathbf{n}}_i=\mathbf{n}_i/b^2$ and $\mathbf{s}_i=[{{r^{-1}_{i}}}, {{{r_{i}}_1^{-1}}},\dots,{{{r_{i}}_k^{-1}}}]^\top$. 

7) SRI SNE \cite{badino2011fast}: Similar to FALS SNE, SRI SNE first transforms the range data from the Cartesian coordinate system to the SCS. $\mathbf{n}_i$ is then obtained by computing the partial derivative of the local tangential surface $s$:
\begin{equation}
	\mathbf{n}_i=\nabla s(\theta_i, \phi_i)=\big[\mathbf{e}_z,\ \mathbf{e}_x,\ \mathbf{e}_y \big]{\mathbf{R}_i}
	\begin{bmatrix}
		1                                                         \\
		\frac{1}{r_i\cos\phi_i}{\partial r_i}/{\partial \theta_i} \\
		\frac{1}{r_i} {\partial r_i}/{\partial \phi_i}
	\end{bmatrix},
	\label{eq.sri1}
\end{equation}
where $\mathbf{R}_i$ is an SO(3) matrix with respect to $\theta_i$ and $\phi_i$.
$\mathbf{e}_z$, $\mathbf{e}_x$ and $\mathbf{e}_y$ are the unit vectors in the $z$, $x$ and $y$ coordinate axes, respectively.
$\nabla s(\theta_i, \phi_i)$ can be obtained by applying standard image convolutional kernels. 

8) LINE-MOD SNE \cite{hinterstoisser2011gradient}: Firstly, the optimal gradient $\nabla z=[{\partial z}/{\partial u}, {\partial z}/{\partial v}]^\top$ of a depth map is computed. Then, a 3D plane is formed by three points $\mathbf{p}_0$, $\mathbf{p}_1$ and $\mathbf{p}_2$:
\begin{equation}
	\begin{aligned}
		\mathbf{p}_0   & =\mathbf{t}(\tilde{\mathbf{p}}_i) z,                                                         \\
		\mathbf{p}_{1} & =\mathbf{t}\left( \tilde{\mathbf{p}}_i+[1,0]^\top  \right)(z+\frac{\partial z}{\partial u}), \\
		\mathbf{p}_{2} & =\mathbf{t}\left( \tilde{\mathbf{p}}_i+[0,1]^\top  \right)(z+\frac{\partial z}{\partial v}),
		\label{eq.line_mod1}
	\end{aligned}
\end{equation}
where $\mathbf{t}(\tilde{\mathbf{p}}_i)$ is the vector along the line of sight that goes through an image pixel $\tilde{\mathbf{p}}_i = [u_i, v_i]^\top$ and is computed using camera intrinsic parameters. The surface normal $\mathbf{n}_i$ can be computed using:
\begin{equation}
	\mathbf{n}_i = \frac{(\mathbf{p}_1 - \mathbf{p}_0) \times (\mathbf{p}_1 - \mathbf{p}_2)}{\|(\mathbf{p}_1 - \mathbf{p}_0) \times (\mathbf{p}_1 - \mathbf{p}_2)\|_2}.
	\label{eq.line_mod2}
\end{equation}

\section{Three-Filters-to-Normal}
\label{sec.methodology}
In this paper, we introduce 3F2N SNE, which is simple to understand and use. Our SNE can compute surface normals from  structured range sensor data using only three filters: 1) a horizontal image gradient filter, 2) a vertical image gradient filter and 3) a mean/median filter.

A 3D point $\mathbf{p}_i=[x,y,z]^\top$ in the CCS can be transformed to $\tilde{\mathbf{p}}_i=[u,v]^\top$ using \cite{hartley2003multiple}:
\begin{equation}
	z\begin{bmatrix}
		u \\v\\1
	\end{bmatrix}=\mathbf{K}\mathbf{p}_i=\begin{bmatrix}
		f_x & 0   & u_\text{o} \\
		0   & f_y & v_\text{o} \\
		0   & 0   & 1
	\end{bmatrix}\begin{bmatrix}
		x \\
		y \\
		z
	\end{bmatrix},
	\label{eq.intrinisc_matrix}
\end{equation}
where $\mathbf{K}$ is the camera intrinsic matrix, $\mathbf{p}_\text{o}=[u_\text{o},v_\text{o}]^\top$ is the image principal point, and $f_x$ and $f_y$ are the camera focal lengths (in pixels) in the $x$ and $y$ directions, respectively. Combining (\ref{eq.plane_function}) and (\ref{eq.intrinisc_matrix}) results in:
\begin{equation}
	\frac{1}{z}=-\frac{1}{b}\bigg(n_x\frac{u-u_\text{o}}{f_x}+n_y\frac{v-v_\text{o}}{f_y}+n_z\bigg).
	\label{eq.1/z_n}
\end{equation}
Differentiating (\ref{eq.1/z_n}) with respect to $u$ and $v$ leads to:
\begin{equation}
	\frac{\partial 1/z}{\partial u}=-\frac{n_x}{b f_x},\ \ \ \ \ \frac{\partial 1/z}{\partial v}=-\frac{n_y}{b f_y},
	\label{eq.d1/z_to_du_dv}
\end{equation}
which can be approximated by respectively performing  horizontal and vertical image gradient filters, \eg, Sobel, Scharr and Prewitt, on the inverse depth image (an image storing the values of $1/z$).
Rearranging (\ref{eq.d1/z_to_du_dv}) results in the following expressions of $n_x$ and $n_y$:
\begin{equation}
	n_x=-b f_x \frac{\partial 1/z}{\partial u}, \ \ \ \ n_y=-b f_y \frac{\partial 1/z}{\partial v}.
	\label{eq.nx1_ny1}
\end{equation}
Given an arbitrary ${\mathbf{q}_{i}}_j\in \mathbf{Q}_i$, we can compute the corresponding ${n_z}_j$ by plugging (\ref{eq.nx1_ny1}) into (\ref{eq.plane_function}):
\begin{equation}
	{n_z}_j=b\frac{ f_x \Delta {x_i}_j \frac{\partial 1/z}{\partial u} + f_y \Delta {y_i}_j \frac{\partial 1/z}{\partial v} }{\Delta {z_i}_j},
	\label{eq.nz1}
\end{equation}
where ${\mathbf{r}_i}_j={\mathbf{q}_i}_j-{\mathbf{p}_i}=[\Delta {x_i}_j, \Delta {y_i}_j, \Delta {z_i}_j]^\top$.
Since (\ref{eq.nx1_ny1}) and (\ref{eq.nz1}) have a common factor of $-b$, they can be simplified as:
\begin{equation}
	\begin{split}
		&{n_x=f_x \frac{\partial 1/z}{\partial u}, \ \ \ \ \ \  n_y=f_y \frac{\partial 1/z}{\partial v}},\\
		&{\hat{n}_z}=-\Phi\Bigg\{\frac{\Delta {x_i}_j n_x + \Delta {y_i}_j n_y }{\Delta {z_i}_j} \Bigg\},\ \  j=1,\dots, k,
		\label{eq.nx2_ny2_nz2}
	\end{split}
\end{equation}
where $\Phi\{\cdot\}$ represents the manner of  $\hat{n}_z$ (the optimum $n_z$) estimation.
In our previous work \cite{fan2020sne}, $\mathbf{n}_i$ is written in spherical coordinates and $\Phi(\cdot)$ is formulated as an energy minimization problem \cite{fan2020sne}, which is computationally intensive. Hence, in this paper, $\Phi\{\cdot\}$ represents a mean/median filtering operation used to estimate $n_z$. Please note: if the depth value of $\mathbf{p}_i$ is identical to those of all its neighboring points ${\mathbf{q}_{i}}_j\in \mathbf{Q}_i$, we consider that the direction of its corresponding surface normal is perpendicular to the image plane and simply set $\mathbf{n}_i$ to $[0, 0, -1]^\top$. The performances of estimating $\mathbf{n}_i$ using the mean filter and using the median filter will be compared in Section \ref{sec.experimental_results}.

\begin{figure*}[!t]
	\centering
	\subfigure[]{
		\includegraphics[width=0.1816\textwidth]{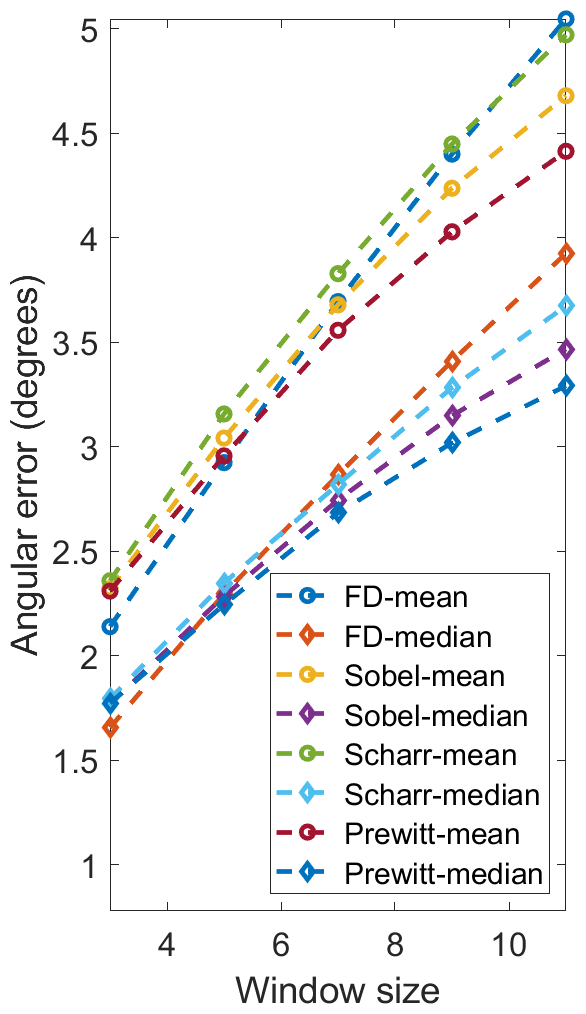}
		\label{fig.easy_line}
	}
	\subfigure[]{
		\includegraphics[width=0.1816\textwidth]{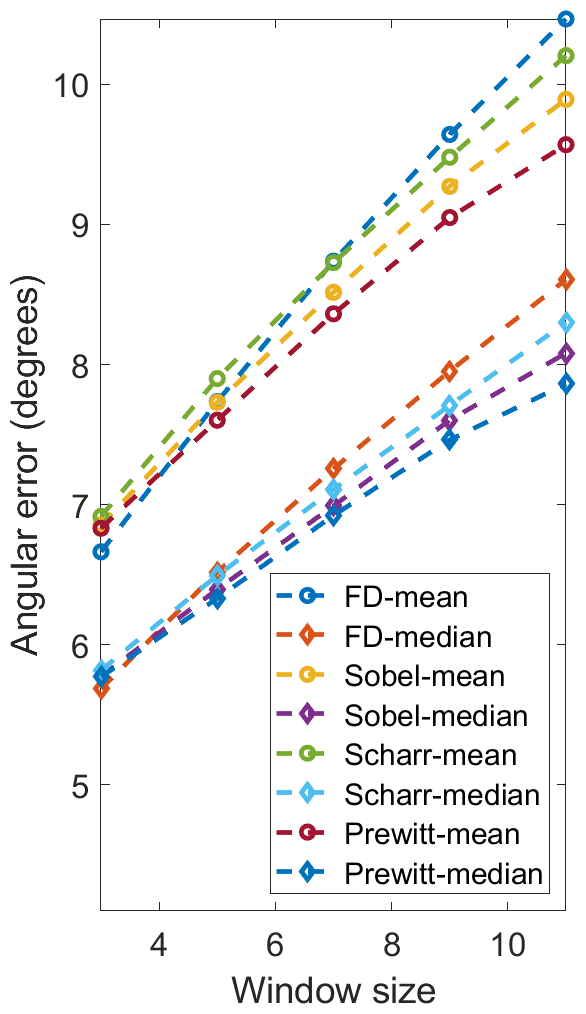}
		\label{fig.medium_line}
	}
	\subfigure[]{
		\includegraphics[width=0.1816\textwidth]{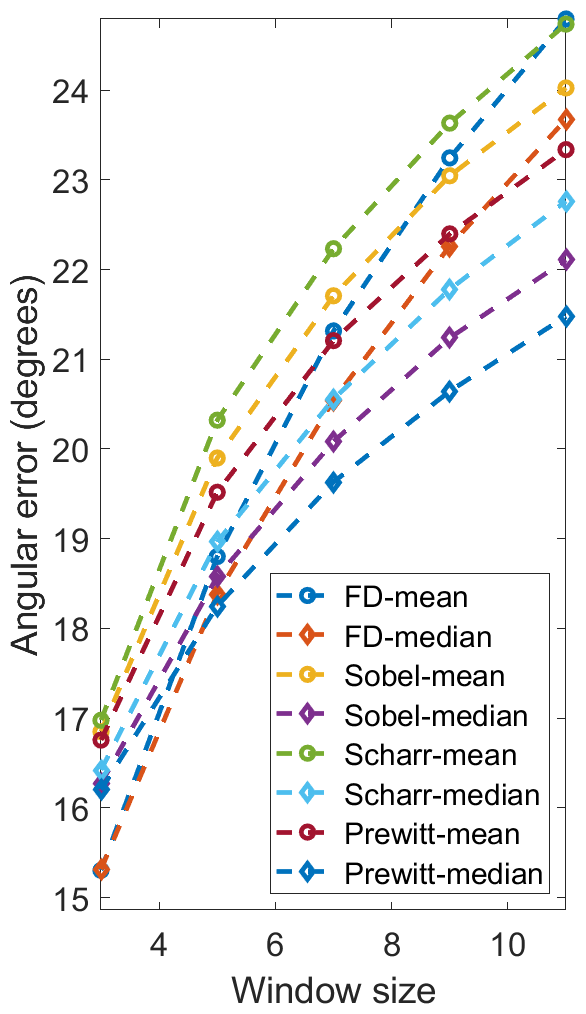}
		\label{fig.hard_line}
	}
	\subfigure[]{
		\includegraphics[width=0.1816\textwidth]{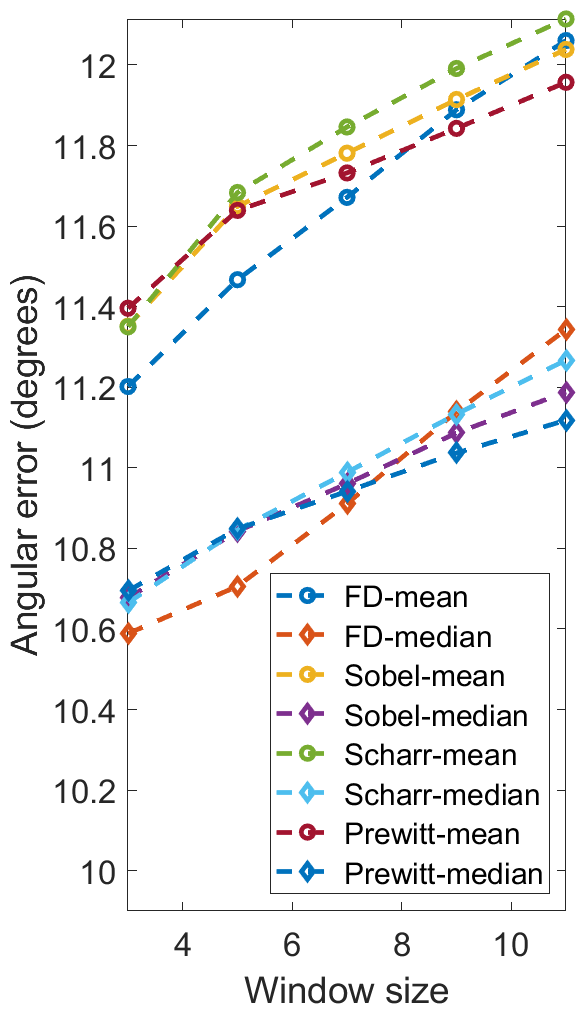}
		\label{fig.indoor_line}
	}
	\subfigure[]{
		\includegraphics[width=0.1816\textwidth]{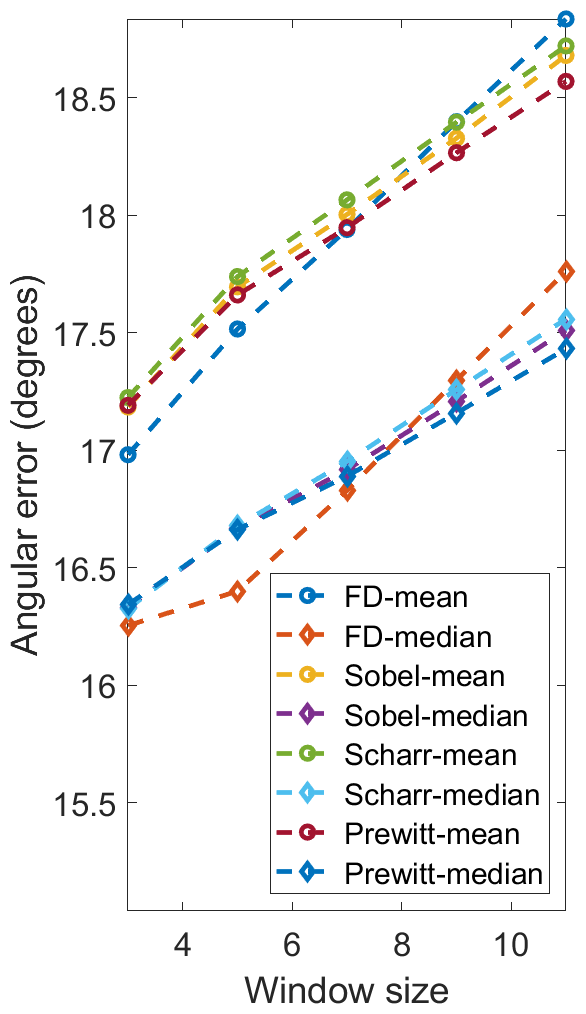}
		\label{fig.outdoor_line}
	}
	\caption{$e_\text{A}$ comparisons with respect to different filter types and sizes: (a) 3F2N-easy dataset; (b) 3F2N-medium dataset; (c) 3F2N-hard dataset; (d) DIODE indoor dataset \cite{diode_dataset}; (e) DIODE  outdoor dataset \cite{diode_dataset}. Please note: (a)-(e) use different scales.}
	\label{fig.line}

\end{figure*}

Specifically, for a stereo camera, $f_x=f_y=f$, and the relationship between depth $z$ and disparity $d$ is as follows \cite{fan2020computer}:
\begin{equation}
	z=\frac{f t_c}{d},
	\label{eq.z_d}
\end{equation}
where $t_c$ is the stereo rig baseline. Therefore,
\begin{equation}
	\begin{aligned}
		\frac{\partial 1/z}{\partial u}=\frac{\partial 1/z}{\partial d}\frac{\partial d}{\partial u}=\frac{1}{f t_c} \frac{\partial d}{\partial u}, \\\frac{\partial 1/z}{\partial v}=\frac{\partial 1/z}{\partial d}\frac{\partial d}{\partial v}=\frac{1}{f t_c} \frac{\partial d}{\partial v}.
	\end{aligned}
	\label{eq.dd_2_dz}
\end{equation}
Plugging (\ref{eq.z_d}) and (\ref{eq.dd_2_dz}) into (\ref{eq.nx2_ny2_nz2}) results in:
\begin{equation}
	\begin{split}
		&{n_x= {\partial d}/{\partial u}, \ \ \ \ \ \ \ \ \  \ \  n_y= {\partial d}/{\partial v}},\\
		&{\hat{n}_z}=-\Phi\Bigg\{ \frac{\Delta {x_i}_j n_x + \Delta {y_i}_j n_y }{\Delta {z_i}_j} \Bigg\},\ j = 1,\dots,k.
		\label{eq.nx2_ny2_nz3}
	\end{split}
\end{equation}
Therefore, our SNE can also estimate surface normals from a disparity image using three filters.

\section{Experiments}
\label{sec.experimental_results}

\begin{figure*}[!t]
	\centering
	\includegraphics[width=0.96\textwidth]{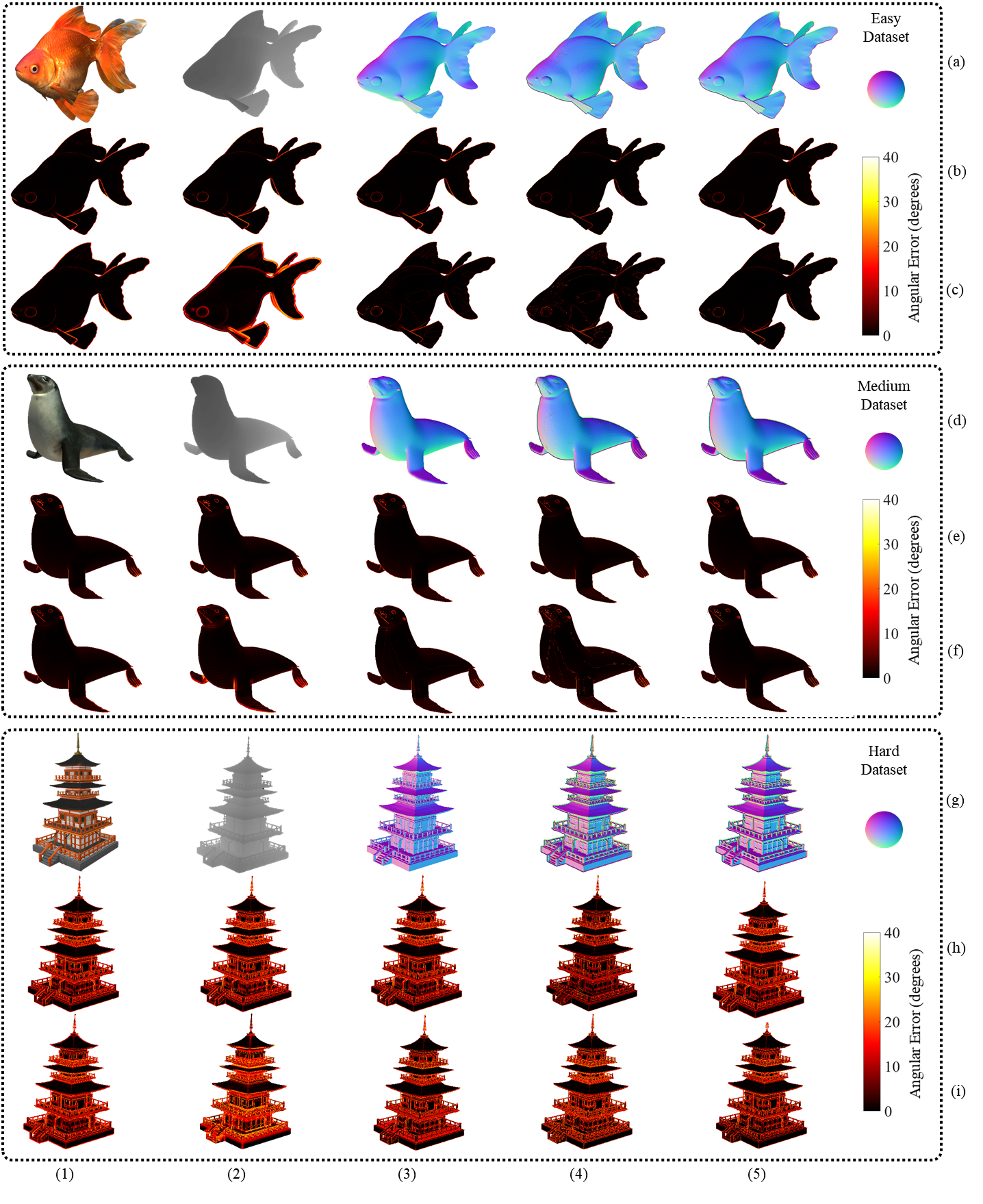}
	\caption{Examples of the experimental results:	(1)--(5) columns on (a), (d) and (g) rows show the 3D mesh models, depth images, surface normal ground truth and the experimental results obtained using FD-Mean and FD-Median SNEs, respectively;  (1)--(5) columns on  (b), (e) and (h) rows show the angular error maps obtained by PlaneSVD/PlanePCA \cite{jordan2014quantitative}, VectorSVD \cite{klasing2009comparison}, AreaWeighted \cite{klasing2009comparison}, AngleWeighted \cite{klasing2009comparison} and FALS \cite{badino2011fast} SNEs, respectively; (1)--(5) columns on  (c), (f) and (i) rows show the angular error maps obtained by SRI \cite{badino2011fast}, LINE-MOD \cite{hinterstoisser2011gradient}, SNE-RoadSeg \cite{fan2020sne}, FD-Mean and FD-Median SNEs, respectively.}
	\label{fig.experimental_results}
\end{figure*}

\subsection{Datasets and Evaluation}
In our experiments, we used 24 3D mesh models from Free3D\footnote{\url{free3d.com}} to create three datasets (eight models in each dataset). According to different difficulty levels, we name our datasets  ``easy'', ``medium'' and ``hard'', respectively. Each 3D mesh model is first fixed at a certain position. A virtual range sensor with pre-set intrinsic parameters is then used to capture depth images at 1800--2500 different view points. At each view point, a $480\times 640$ pixel depth image is generated by rendering the 3D mesh model using OpenGL Shading Language\footnote{\url{opengl.org/sdk/docs/tutorials/ClockworkCoders/glsl_overview.php}} (GLSL). However, since the OpenGL rendering process applies linear interpolation by default, rendering surface normal images is infeasible. Hence, the surface normal of each triangle, constructed by three mesh vertices, is considered to be the ground truth surface normal of any 3D points residing on this triangle.
Our datasets are publicly available at \url{sites.google.com/view/3f2n/datasets} for research purposes. 
In addition to our datasets, we also utilize two real-world datasets: 1) the DIODE dataset\footnote{\url{diode-dataset.org}} \cite{diode_dataset} and 2) the ScanNet\footnote{\url{www.scan-net.org/}} \cite{dai2017scannet} dataset to evaluate the SNE performance on noisy depth data.
Furthermore, we utilize two metrics: a) the average angular error (AAE) $e_\text{A}$  and b) the proportion of good pixels (PGP) $e_\text{P}$ \cite{lu2017symps}:
\begin{equation}
	e_\text{A} =
	\frac{1}{m}\sum_{k=1}^{m}
	\psi_k
	, \ \ \ \ \ \ \ e_\text{P}(\varphi)=\frac{1}{m}\sum_{k=1}^{m} \delta
	(\psi_k, \varphi)
	\label{eq.ea_ep}
\end{equation}
to quantify the SNE accuracy,  where:
\begin{equation}
	\delta(\psi_k, \varphi)=\left\{
	\begin{array}{lr}
		0 & (\psi_k>\varphi)    \\
		1 & (\psi_k\leq\varphi)
	\end{array},
	\right.
\end{equation}
\begin{equation}
	\psi_k=\cos^{-1}\bigg(
	\frac{
		\langle{
			\mathbf{n}_k, {\hat{\mathbf{n}}_k}
			\rangle}
	}
	{
		\|\mathbf{n}_k\|_2\|\hat{\mathbf{n}}_k\|_2
	}
	\bigg),
\end{equation}
$m$ is the number of 3D points used for evaluation, $\varphi$ is the angular error tolerance, and $\mathbf{n}_k$ and $\hat{\mathbf{n}}_k$ are the estimated and ground truth surface normals, respectively. In addition to accuracy, we also record the SNE processing time $t$ (ms) and introduce a new metric:
\begin{equation}
	\pi= e_\text{A} t\ (\text{degrees}/\text{kHz})
\end{equation}
to quantify the trade-off between the speed and accuracy of a given SNE. A fast and precise SNE achieves a low $\pi$ score.

\begin{table}[!t]
	\begin{center}
		\setlength\arrayrulewidth{0.8pt}
		\footnotesize
		\caption{The runtime (ms) of the CPU implementations (using a single thread) with respect to different image gradient filters and mean/median filters. }
		\label{tab.cpu_runtime}
		\begin{tabular}{l|c|c}
			\hline
			Gradient filter & Mean filter    & Median filter   \\
			\hline
			FD              & \textbf{3.722} & \textbf{10.973} \\
			Sobel           & 3.824          & 11.167          \\
			Scharr          & 3.848          & 11.355          \\
			Prewitt         & 3.743          & 11.065          \\
			\hline
		\end{tabular}
	\end{center}
\end{table}

\begin{table}[!t]
	\begin{center}
		\setlength\arrayrulewidth{0.8pt}
		\footnotesize
		\caption{The runtime (ms) of the GPU implementations with respect to different image gradient filters and mean/median filters. }
		\label{table.gpu_runtime}
		\begin{tabular}{l|c|c|c}
			\hline
			Method         & Jetson TX2        & GTX 1080 Ti       & RTX 2080 Ti       \\
			\hline
			
			FD-Mean        & \textbf{0.823521} & \textbf{0.049504} & \textbf{0.046944} \\
			Sobel-Mean     & 0.855843          & 0.052288          & 0.051232          \\
			Scharr-Mean    & 0.860319          & 0.052320          & 0.051280          \\
			Prewitt-Mean   & 0.857762          & 0.052256          & 0.050816          \\
			
			\hline
			FD-Median      & \textbf{1.206337} & \textbf{0.102368} & \textbf{0.065536} \\
			Sobel-Median   & 1.217023          & 0.104608          & 0.067840          \\
			Scharr-Median  & 1.239041          & 0.105376          & 0.071008          \\
			Prewitt-Median & 1.240479          & 0.105152          & 0.069024          \\
			
			\hline
		\end{tabular}
	\end{center}
\end{table}

\begin{table*}[!t]
	\begin{center}
		\vspace{0in}
		\footnotesize
		\caption{Comparisons among geometry-based SNEs on our created synthetic datasets.}
		\label{table.ten_methods_t_pi}
		\setlength\arrayrulewidth{0.7pt}
		\begin{tabular}{l|r|r|r|r|r|r|r}
			\hline
			\multicolumn{1}{c|}{\multirow{2}*{Method}} & \multicolumn{1}{c|}{\multirow{2}{*}{$t$ (ms) $\downarrow$}} & \multicolumn{3}{c|}{$e_\text{A}$ (degrees)  $\downarrow$} & \multicolumn{3}{c}{$\pi\ $ (degrees/kHz)  $\downarrow$}                                                                                                                  \\
			\cline{3-8}
			&                                                             & \multicolumn{1}{c|}{Easy}                                 & \multicolumn{1}{c|}{Medium}                             & \multicolumn{1}{c|}{Hard} & \multicolumn{1}{c|}{Easy} & \multicolumn{1}{c|}{Medium} & \multicolumn{1}{c}{Hard} \\
			\hline
			PlaneSVD \cite{klasing2009realtime}        & 393.69                                                      & 2.07                                                      & 6.07                                                    & 17.59                     & 813.87                    & 2389.73                     & 6923.18
			\\

			PlanePCA \cite{jordan2014quantitative}     & 631.88                                                      & 2.07                                                      & 6.07                                                    & 17.59                     & 1306.29                   & 3835.59                     & 11111.92
			\\
			
			VectorSVD \cite{klasing2009comparison}     & 563.21                                                      & 2.13                                                      & 6.27                                                    & 18.01                     & 1199.63                   & 3529.11                     & 10142.34
			\\
			AreaWeighted \cite{klasing2009comparison}  & 1092.24                                                     & 2.20                                                      & 6.27                                                    & 17.03                     & 2407.74                   & 6843.56                     & 18600.68
			\\
			AngleWeighted \cite{klasing2009comparison} & 1032.88                                                     & 1.79                                                      & \textbf{5.67}                                           & \textbf{13.26}            & 1850.00                   & 5855.62                     & 13693.24
			\\
			FALS \cite{badino2011fast}                 & 4.11                                                        & 2.26                                                      & 6.14                                                    & 17.34                     & 9.26                      & 25.20                       & 71.17
			\\
			SRI \cite{badino2011fast}                  & 12.18                                                       & 2.64                                                      & 6.71                                                    & 19.61                     & 32.18                     & 81.66                       & 238.78
			\\
			LINE-MOD \cite{hinterstoisser2011gradient} & 6.43                                                        & 6.53                                                      & 9.94                                                    & 31.45                     & 41.93                     & 63.84                       & 202.08
			\\
			SNE-RoadSeg \cite{fan2020sne}              & 7.92                                                        & 2.04                                                      & 6.28                                                    & 16.37                     & 16.16                     & 49.74                       & 129.65 
			\\
			
			\hline
			FD-Mean (ours)                             & \textbf{3.72}                                               & 2.14                                                      & 6.66                                                    & 15.30                     & \textbf{7.96}             & \textbf{24.80}              & \textbf{56.96}
			\\
			
			FD-Median (ours)                           & 10.97                                                       & \textbf{1.66}                                             & 5.69                                                    & 15.31                     & 18.18                     & 62.38                       & 168.03  
			\\
			
			\hline
		\end{tabular}
	\end{center}
\end{table*}

\begin{table*}[!t]
	\begin{center}
		\vspace{0in}
		\footnotesize
		\caption{$e_\text{P}$ comparison among geometry-based SNEs with respect to different $\varphi$ on our created synthetic datasets.}
		\label{table.ten_methods}
		\setlength\arrayrulewidth{0.7pt}
		\begin{tabular}{l|ccc|ccc|ccc}
			\hline
			\multicolumn{1}{c|}{\multirow{3}*{Method} } & \multicolumn{9}{c}{$e_\text{P}$  $\uparrow$}                                                                                                                                                                                                    \\
			\cline{2-10}
			& \multicolumn{3}{c|}{Easy}                    & \multicolumn{3}{c|}{Medium} & \multicolumn{3}{c}{Hard}                                                                                                                                           \\
			\cline{2-10}
			
			& $\varphi$=10$^\circ$                         & $\varphi$=20$^\circ$        & $\varphi$=30$^\circ$     & $\varphi$=10$^\circ$ & $\varphi$=20$^\circ$ & $\varphi$=30$^\circ$ & $\varphi$=10$^\circ$ & $\varphi$=20$^\circ$ & $\varphi$=30$^\circ$ \\
			\hline
			
			PlaneSVD \cite{klasing2009realtime}         & 0.9648                                       & 0.9792                      & 0.9855                   & 
			0.8621                                      & 
			0.9531                                      & 
			0.9718                                      & 
			0.6202                                      & 
			0.7394                                      & 
			0.7914

			\\
			PlanePCA \cite{jordan2014quantitative}      & 
			0.9648                                      & 
			0.9792                                      & 
			0.9855                                      & 
			0.8621                                      & 
			0.9531                                      & 
			0.9718                                      & 
			0.6202                                      & 
			0.7394                                      & 
			0.7914

			\\			
			VectorSVD \cite{klasing2009comparison}      & 0.9643                                       & 0.9777                      & 0.9846                   & 
			
			0.8601                                      & 
			0.9495                                      & 
			0.9683                                      & 
			0.6187                                      & 
			0.7346                                      & 
			0.7848
			\\
			AreaWeighted \cite{klasing2009comparison}   & 0.9636                                       & 0.9753                      & 0.9819                   & 
			
			0.8634                                      & 
			0.9504                                      & 
			0.9665                                      & 
			0.6248                                      & 
			0.7448                                      & 
			0.7977
			\\
			AngleWeighted \cite{klasing2009comparison}  & \textbf{0.9762}                              & \textbf{0.9862}             & \textbf{0.9893}          & 
			\textbf{0.8814}                             & 
			\textbf{0.9711}                             & 
			\textbf{0.9809}                             & 
			0.6625                                      & 
			\textbf{0.8075}                             & 
			\textbf{0.8651}
			\\
			FALS \cite{badino2011fast}                  & 0.9654                                       & 0.9794                      & 0.9857                   & 
			
			0.8621                                      & 
			0.9547                                      & 
			0.9731                                      & 
			0.6209                                      & 
			0.7433                                      & 
			0.7961
			\\
			SRI \cite{badino2011fast}                   & 0.9499                                       & 0.9713                      & 0.9798                   & 
			0.8431                                      & 
			0.9403                                      & 
			0.9633                                      & 
			0.5594                                      & 
			0.6932                                      & 
			0.7605
			\\
			LINE-MOD \cite{hinterstoisser2011gradient}  & 
			0.8542                                      & 0.9085                                       & 0.9343                      & 
			0.7277                                      & 
			0.8803                                      & 
			0.9282                                      & 
			0.3375                                      & 
			0.4757                                      & 
			0.5636
			\\
			
			SNE-RoadSeg \cite{fan2020sne}               & 0.9693                                       & 0.9810                      & 0.9871                   & 0.8618               & 0.9512               & 0.9725               & 0.6226               & 0.7589               & 0.8113
			\\

			\hline
			FD-Mean (ours)                              & 0.9563                                       & 0.9767                      & 0.9864                   & 
			0.8349                                      & 
			0.9423                                      & 
			0.9674                                      & 
			0.6191                                      & 
			0.7671                                      & 
			0.8368
			\\
			FD-Median (ours)                            & 0.9723                                       & 0.9829                      & 
			0.9889                                      & 
			0.8722                                      & 
			0.9600                                      & 
			0.9766                                      & 
			\textbf{0.6631}                             & 
			0.7821                                      & 
			0.8289
			\\
			\hline
		\end{tabular}
	\end{center}
	\vspace{-1.5em}
\end{table*}

\begin{table*}[!t]
	\begin{center}
		\vspace{0in}
		\footnotesize
		\caption{Comparisons among different geometry-based SNEs on the DIODE dataset \cite{diode_dataset}.}
		\label{table.diode_methods}
		\setlength\arrayrulewidth{0.7pt}
		\begin{tabular}{l|r|r|r|r|r|r|r|r|r|r|r}
			\hline
			\multicolumn{1}{c|}{\multirow{3}*{Method}} & \multicolumn{1}{c|}{\multirow{3}*{$t$ (ms) $\downarrow$}} & \multicolumn{2}{c|}{$e_\text{A}$ (degrees) $\downarrow$} & \multicolumn{2}{c|}{$\pi\ $ (degrees/kHz) $\downarrow$} & \multicolumn{6}{c}{$e_\text{p}$ $\uparrow$}                                                                                                                                                                                                                                      \\
			\cline{3-12}
			
			&                                                           & \multicolumn{1}{c|}{\multirow{2}*{ Indoor}}              & \multicolumn{1}{c|}{\multirow{2}*{ Outdoor}}            & \multicolumn{1}{c|}{\multirow{2}*{ Indoor}} & \multicolumn{1}{c|}{\multirow{2}*{ Outdoor}} & \multicolumn{3}{c|}{\multirow{1}*{Indoor}} & \multicolumn{3}{c}{\multirow{1}*{Outdoor}}                                                                                             \\
			\cline{7-12}
			
			&                                                           &                                                          &                                                         &                                             &                                              & $\varphi$=10$^\circ$                       & $\varphi$=20$^\circ$                       & $\varphi$=30$^\circ$ & $\varphi$=10$^\circ$ & $\varphi$=20$^\circ$ & $\varphi$=30$^\circ$ \\
			\hline

			PlaneSVD \cite{klasing2009realtime}        & 883.458                                                   & 10.888                                                   & 16.579                                                  & 9619.002                                    & 14646.762                                    & 0.693                                      & 0.924                                      & 0.942                & 0.574                & 0.763                & 0.811

			\\		
			
			PlanePCA \cite{jordan2014quantitative}     & 1501.707                                                  & 10.888                                                   & 16.579                                                  & 16350.436                                   & 24896.650                                    & 0.693                                      & 0.924                                      & 0.942                & 0.574                & 0.763                & 0.811

			\\

			VectorSVD \cite{klasing2009comparison}     & 1327.847                                                  & 10.868                                                   & 16.514                                                  & 14431.572                                   & 21928.464                                    & 0.696                                      & 0.925                                      & 0.942                & 0.577                & \textbf{0.764}       & 0.812

			\\

			AreaWeighted \cite{klasing2009comparison}  & 2522.729                                                  & 10.887                                                   & 16.560                                                  & 27465.203                                   & 41775.635                                    & 0.691                                      & 0.924                                      & 0.942                & 0.572                & 0.763                & 0.812

			\\
			
			AngleWeighted \cite{klasing2009comparison} & 2661.607                                                  & 10.759                                                   & 16.545                                                  & 28636.496                                   & 44037.086                                    & {0.689}                                    & \textbf{0.925}                             & \textbf{0.943}       & {0.568}              & {0.763}              & \textbf{0.815}

			\\
			FALS \cite{badino2011fast}                 & 10.706                                                    & 11.072                                                   & 16.671                                                  & 118.531                                     & 178.474                                      & 0.682                                      & 0.923                                      & 0.941                & 0.571                & 0.759                & 0.813

			\\

			SRI \cite{badino2011fast}                  & 39.075                                                    & 11.154                                                   & 16.903                                                  & 435.854                                     & 660.481                                      & 0.685                                      & 0.918                                      & 0.936                & 0.571                & 0.757                & 0.807

			\\

			LINE-MOD \cite{hinterstoisser2011gradient} & 17.026                                                    & 12.839                                                   & 17.272                                                  & 218.593                                     & 294.071                                      & 0.663                                      & 0.886                                      & 0.907                & 0.577                & 0.749                & 0.796

			\\
			
			SNE-RoadSeg \cite{fan2020sne}              & 20.310                                                    & \textbf{10.316}                                          & \textbf{15.431}                                         & 209.599                                     & 313.383                                      & 0.692                                      & 0.921                                      & 0.941                & 0.555                & 0.760                & 0.810

			\\

			\hline
			FD-Mean (ours)
			& \textbf{9.511}                                            & 11.202                                                   & 16.981                                                  & \textbf{106.540}                            & \textbf{161.507}                             & 0.613                                      & 0.854                                      & 0.903                & 0.477                & 0.713                & 0.779

			\\

			FD-Median (ours)
			& 30.193                                                    & {10.589}                                                 & {16.254}                                                & {319.705}                                   & {490.769}                                    & \textbf{0.706}                             & 0.922                                      & 0.940                & \textbf{0.578}       & 0.761                & 0.809

			\\

			\hline
		\end{tabular}
	\end{center}
\end{table*}

\begin{table*}[!t]
	\begin{center}
		\vspace{0in}
		\footnotesize
		\caption{Comparisons among different geometry-based SNEs on the ScanNet dataset \cite{dai2017scannet}.}
		\label{table.scannet}
		\setlength\arrayrulewidth{0.7pt}
		\begin{tabular}{l|r|r|r|r|r|r}
			\hline
			\multicolumn{1}{c|}{\multirow{2}*{Method}} & \multicolumn{1}{c|}{\multirow{2}{*}{$t$ (ms) $\downarrow$}} & \multicolumn{1}{c|}{\multirow{2}*{$e_\text{A}$ (degrees) $\downarrow$}} & \multicolumn{1}{c|}{\multirow{2}*{$\pi\ $ (degrees/kHz)  $\downarrow$}} & \multicolumn{3}{c}{$e_\text{p}$ $\uparrow$}                                               \\
			\cline{5-7}
			&                                                             &                                                                         &                                                                         & $\varphi$=10$^\circ$                        & $\varphi$=20$^\circ$ & $\varphi$=30$^\circ$ \\
			\hline
			PlaneSVD \cite{klasing2009realtime}        & 462.349                                                     & 13.164                                                                  & 6086.362                                                                & 0.645                                       & 0.861                & 0.890
			\\

			PlanePCA \cite{jordan2014quantitative}     & 782.475                                                     & 13.164                                                                  & 10300.501                                                               & 0.645                                       & 0.861                & 0.890
			\\
			
			VectorSVD \cite{klasing2009comparison}     & 687.917                                                     & 13.239                                                                  & 9107.333                                                                & 0.646                                       & 0.856                & 0.887
			\\
			AreaWeighted \cite{klasing2009comparison}  & 1391.188                                                    & 13.213                                                                  & 18381.767                                                               & 0.641                                       & 0.858                & 0.889
			\\
			AngleWeighted \cite{klasing2009comparison} & 1475.558                                                    & 12.958                                                                  & 19120.281                                                               & 0.642                                       & 0.863                & \textbf{0.894}
			\\
			FALS \cite{badino2011fast}                 & 5.308                                                       & 13.256                                                                  & 70.363                                                                  & 0.639                                       & 0.860                & 0.891
			\\
			SRI \cite{badino2011fast}                  & 15.704                                                      & 13.626                                                                  & 213.983                                                                 & 0.637                                       & 0.849                & 0.881
			\\
			LINE-MOD \cite{hinterstoisser2011gradient} & 7.679                                                       & 14.479                                                                  & 111.184                                                                 & 0.631                                       & 0.834                & 0.866
			\\
			SNE-RoadSeg \cite{fan2020sne}              & 10.634                                                      & 12.669                                                                  & 134.722                                                                 & 0.630                                       & 0.847                & 0.881
			\\
			
			\hline
			FD-Mean (ours)                             & \textbf{4.630}                                              & 13.225                                                                  & \textbf{61.232}                                                         & 0.565                                       & 0.805                & 0.865
			\\
			
			FD-Median (ours)                           & 13.924                                                      & \textbf{12.628}                                                         & 175.832                                                                 & \textbf{0.651}                              & \textbf{0.864}       & 0.893
			\\
			
			\hline
		\end{tabular}
	\end{center}
	\vspace{-2em}
\end{table*}

\begin{table*}[!h]
	\begin{center}
		\vspace{0in}
		\footnotesize
		\caption{$e_\text{A}$ comparison among geometry-based SNEs with respect to different noise levels on our created synthetic datasets.}
		\label{table.noise_level}
		\setlength\arrayrulewidth{0.7pt}
		\begin{tabular}{l|ccc|ccc|ccc}
			\hline
			\multicolumn{1}{c|}{\multirow{3}*{Method} } & \multicolumn{9}{c}{$e_\text{A}$ (degrees)  $\downarrow$}                                                                                                                               \\
			\cline{2-10}
			& \multicolumn{3}{c|}{Low noise level}           & \multicolumn{3}{c|}{Medium noise level} & \multicolumn{3}{c}{High noise level}                                                  \\
			\cline{2-10}
			
			& Easy                                           & Medium                               & Hard                                & Easy & Medium & Hard  & Easy  & Medium & Hard  \\
			\hline
			
			PlaneSVD \cite{klasing2009realtime}         & 2.26                                           & 6.69                                 & 19.45                               & 3.11 & 9.21   & 26.39 & 4.19  & 12.14  & 35.28

			\\
			PlanePCA \cite{jordan2014quantitative}      & 2.26                                           & 6.69                                 & 19.45                               & 3.11 & 9.21   & 26.39 & 4.19  & 12.14  & 35.28

			\\
			VectorSVD \cite{klasing2009comparison}      & 2.34                                           & 6.92                                 & 19.80                               & 3.23 & 9.41   & 27.06 & 4.23  & 12.58  & 36.02
			\\
			AreaWeighted \cite{klasing2009comparison}   & 2.42                                           & 6.86                                 & 18.71                               & 3.36 & 9.43   & 25.59 & 4.36  & 12.56  & 34.06
			\\
			AngleWeighted \cite{klasing2009comparison}  & 1.93                                           & \textbf{6.24}                                 & \textbf{15.69}                               & 2.67 & \textbf{8.51}   & \textbf{21.89} & 3.68  & \textbf{11.34}  & \textbf{28.52}
			\\
			FALS \cite{badino2011fast}                  & 2.49                                           & 6.65                                 & 19.17                               & 3.39 & 9.31   & 26.01 & 4.53  & 12.28  & 34.78
			\\
			SRI \cite{badino2011fast}                   & 2.93                                           & 7.40                                 & 21.59                               & 3.96 & 10.07  & 29.52 & 5.26  & 13.42  & 39.12
			\\
			LINE-MOD \cite{hinterstoisser2011gradient}  & 6.98                                           & 10.93                                & 33.60                               & 8.90 & 14.91  & 39.18 & 12.06 & 17.88  & 55.90
			\\
			
			SNE-RoadSeg \cite{fan2020sne}               & 2.26                                           & 6.91                                 & 18.11                               & 3.05 & 9.42   & 24.66 & 4.09  & 12.56  & 32.64
			\\

			\hline
			FD-Mean (ours)                              & 2.37                                           & 7.32                                 & 16.83                               & 3.23 & 9.97   & 22.92 & 4.27  & 13.32  & 30.51
			\\
			FD-Median (ours)                            & \textbf{1.82}                                           & 6.26                                 & 16.85                               & \textbf{2.49} & 8.52   & 22.96 & \textbf{3.35}  & 11.38  & 30.61
			\\
			\hline
		\end{tabular}
	\end{center}
	
\end{table*}

\begin{figure*}[!t]
	\centering
	\includegraphics[width=0.999999\textwidth]{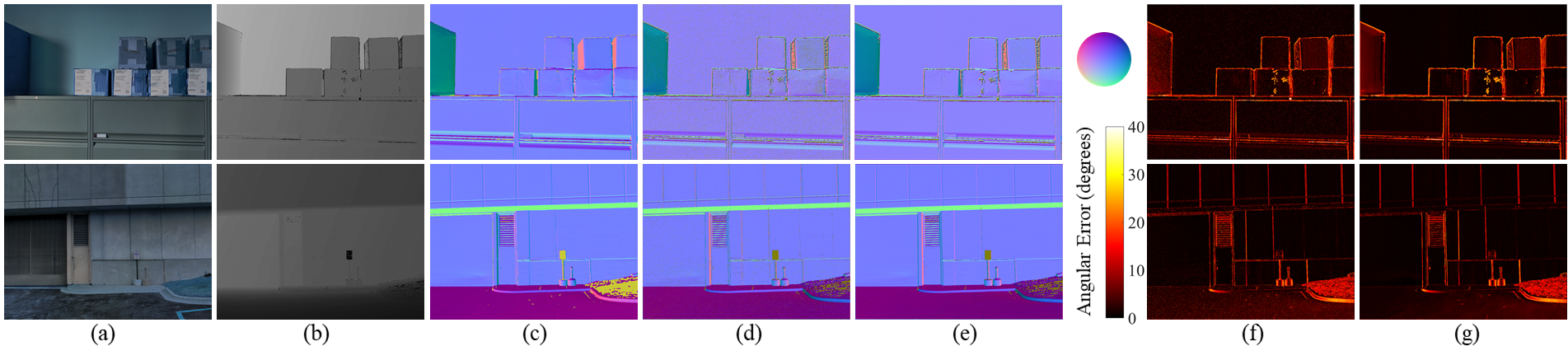}
	\caption{Examples of the DIODE dataset \cite{diode_dataset}: (a) RGB images; (b) depth images; (c) surface normal ground truth; (d) FD-Mean SNE results; (e) FD-Median SNE results; (f) FD-Mean SNE error maps; (g) FD-Median SNE error maps. }
	\label{fig.diode_res}
\end{figure*}

\begin{figure*}[!t]
	\centering
	\includegraphics[width=0.90\textwidth]{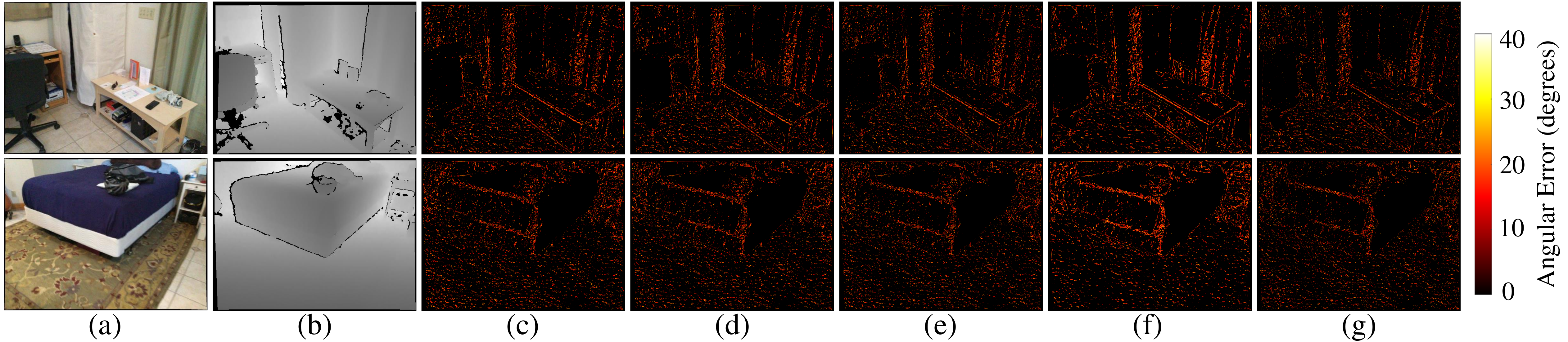}
	\caption{Examples of the ScanNet dataset \cite{dai2017scannet}: (a) RGB images; (b) depth images; (c) PlaneSVD/PlanePCA SNE error maps; (d) AngleWeighted error maps; (e) SNE-RoadSeg SNE error maps; (f) FD-Mean SNE error maps; (g) FD-Median SNE error maps. }
	\label{fig.scannet_res}
\end{figure*}

\subsection{Filter Settings and Implementation Details}
As discussed in Section \ref{sec.methodology}, $n_x$ and $n_y$ can be estimated by convolving an inverse depth image or a disparity map with image convolutional kernels, \eg, Sobel, Scharr, Prewitt, \etc. Hence, in our experiments, we first compare the  accuracy of the surface normals  estimated using the aforementioned convolutional kernels. The brute-force search strategy is then applied to find the best parameters for a $3\times3$ kernel. Our experiments illustrate that finite difference (FD) kernel, \ie, $[-1, 0, 1]$, can achieve the best overall performance.

We implement the proposed SNE in Matlab C and C++ on a CPU and in CUDA on a GPU. The source code is available at \url{github.com/ruirangerfan/three_filters_to_normal}. Similar to the FALS, SRI and LINE-MOD SNE implementations provided in the opencv\_contrib repository,\footnote{\url{github.com/opencv/opencv_contrib}} we use advanced vector extensions 2 (AVX2) and streaming SIMD (single instruction, multiple data) extensions (SSE) instruction sets to optimize our C++ implementation. Since our approach estimates surface normals from an 8-connected neighborhood, we also use memory alignment strategies to speed up our SNE. In the GPU implementation, we first create a texture object in the GPU texture memory and then bind this object with the address of the input depth/disparity image, which greatly reduces the memory requests from the GPU global memory.

\subsection{Performance Evaluation}

We first compare the performances of the proposed SNE with respect to different image gradient filters (FD, Sobel, Scharr and Prewitt) and mean/median filter. $e_\text{A}$ scores achieved on our and the DIODE \cite{diode_dataset} datasets are given in Fig. \ref{fig.line}. The runtime of our implementations on an Intel Core i7-8700K CPU (using a single thread) and three state-of-the-art GPUs (Jetson TX2, GTX 1080 Ti and RTX 2080 Ti) is also given in Table \ref{tab.cpu_runtime} and \ref{table.gpu_runtime}, respectively. We can observe that FD outperforms Sobel, Scharr and Prewitt in terms of $e_\text{A}$ on all datasets. Also, using the median filter can achieve better surface normal accuracy than using the mean filter, because an $n_z$ candidate in (\ref{eq.nz1}) can differ significantly from the ground truth value, introducing significant noise to the mean filter. 
The $e_\text{A}$ scores achieved using FD-Median SNE are lower than those achieved by FD-Mean SNE by $0.5^\circ$, $1.0^\circ$, $0.6^\circ$, $0.7^\circ$ and $0.6^\circ$ with respect to 3F2N-easy, 3F2N-medium, DIODE-indoor \cite{diode_dataset}, DIODE-outdoor \cite{diode_dataset}, and ScanNet \cite{dai2017scannet} datasets, respectively.
However, median filter is much more computationally intensive and time-consuming than the mean filter, because it needs to sort eight $n_z$ candidates  and find the median value. From Table \ref{tab.cpu_runtime} and \ref{table.gpu_runtime}, we can observe that both FD-Mean SNE and FD-Median SNE perform much faster than real-time across different computing platforms. The processing speed of FD-Mean SNE is over 1 kHz and 21 kHz on the Jetson TX2 GPU and RTX 2080 Ti GPU, respectively. Furthermore, FD-Mean SNE performs around 1.4 to 2.1 times faster than the FD-Median SNE. Therefore, the latter achieves the best surface normal accuracy, while the former achieves the best processing speed.

Moreover, we compare 3F2N SNE with all other state-of-the-art geometry-based SNEs, as mentioned in Section \ref{sec.related_work}. Some examples of the experimental results are shown in Fig. \ref{fig.experimental_results}, where it can be seen that the bad estimates mainly reside on the object edges.
Additionally, Table \ref{table.ten_methods_t_pi} shows comparisons of $e_\text{A}$ on the easy, medium and hard datasets,
where we can find that FD-Median SNE achieves the best $e_\text{A}$ score on the easy dataset, while AngleWeighted \cite{klasing2009comparison} SNE achieves the best $e_\text{A}$ scores on the medium and hard datasets.
Meanwhile, the $e_\text{A}$ scores achieved by FD-Median SNE and AngleWeighted \cite{klasing2009comparison} SNE are very similar. The runtime (C++ implementations using a single thread) and $\pi$ scores  achieved  by the aforementioned SNEs are given in Table \ref{table.ten_methods_t_pi}, where we can observe that the averaging-based SNEs are the most time-consuming ones, while FD-Mean SNE achieves the fastest processing speed. Furthermore, FD-Mean, FALS \cite{badino2011fast} and FD-Median SNEs occupy the first three places, respectively, in terms of $\pi$ score. Moreover, Table \ref{table.ten_methods} compares their PGP scores with respect to different $\varphi$ on the easy, medium and hard datasets, where we can see that AngleWeighted \cite{klasing2009comparison} SNE achieves the best  $e_\text{P}$ scores,  except for $\varphi=10^\circ$ (hard dataset). However, according to Table \ref{table.ten_methods_t_pi}, AngleWeighted \cite{klasing2009comparison} SNE is extremely time-consuming and achieves a very bad $\pi$ score. On the other hand, FD-Median SNE and AngleWeighted \cite{klasing2009comparison} SNE achieve similar $e_\text{P}$ scores, but the former performs about 100 times faster than the latter. Furthermore, we add random Gaussian noise to our created depth images and provide comprehensive comparisons of these SNEs in the supplement. 

In addition to our created datasets, we also use the DIODE \cite{diode_dataset} and ScanNet \cite{dai2017scannet} datasets, respectively, to compare the performances of the above-mentioned SNEs on noisy depth data. Examples of our experimental results are shown in Figs. \ref{fig.diode_res} and \ref{fig.scannet_res}, respectively. The runtime and average angular errors obtained by different SNEs are given in Tabs \ref{table.diode_methods} and \ref{table.scannet}, respectively, where it can be seen that FD-Mean SNE is the fastest among all SNEs, while FD-Median SNE achieves the lowest $e_\text{p}$ when $\varphi=10^\circ$. FD-Mean greatly minimizes the trade-off between speed and accuracy. Therefore,  3F2N SNE outperforms all other state-of-the-art geometry-based SNEs in terms of both accuracy and speed. Researchers can use either FD-Mean SNE or FD-Median SNE in their work, according to their demand for speed or accuracy.

\section{Discussion}
\label{sec.discussion}

An SNE can be applied in a variety of computer vision and robotics tasks. 
In this paper, we perform ElasticFusion \cite{whelan2015elasticfusion}, a real-time dense visual simultaneous localization and mapping (SLAM) algorithm, on the ICL-NUIM RGB-D dataset \cite{handa:etal:ICRA2014} with and without surface normal information incorporated, respectively. According to the quantitative analysis of our experimental results, the 3D geometry reconstruction accuracy can be improved by approximately 19\%, when using the surface normal information obtained by 3F2N SNE. Examples of the experimental results are given in the supplement.

Moreover, we have also proven in \cite{fan2020sne} and \cite{wang2021tcyb} that surface normal information can be employed for various planar surface segmentation applications. Therefore, we believe that 3F2N SNE can be utilized to extract informative features for CNNs in various autonomous driving perception tasks, without affecting their training/prediction speed. 

Finally, it is emphasized that the proposed SNE is essentially different from the approaches developed for dominant surface normal estimation (or Manhattan frame model inference \cite{straub2017manhattan}). The latter aims at estimating the surface normal of each planar surface instead of the one of every single pixel.

\section{Conclusion}
\label{sec.conclusion_future_work}
In this paper, we presented a precise and ultrafast SNE named 3F2N for structured range data. Our proposed SNE can compute surface normals from an inverse depth image or a disparity image using three filters, namely, a horizontal image gradient filter, a vertical image gradient filter and a mean/median filter. To evaluate the performance of our proposed SNE, we created three datasets (containing about 60k pairs of depth images and the corresponding surface normal ground truth) using 24 3D mesh models. Our datasets are also publicly available for research purposes. According to our experimental results, FD outperforms other image gradient filters, \eg, Sobel, Scharr and Prewitt, in terms of both precision and speed. FD-Median SNE achieves the best surface normal precision ($1.66^\circ$, $5.69^\circ$ and $15.31^\circ$ on easy, medium and hard datasets, respectively), while FD-Mean SNE is most effective for minimizing the trade-off between speed and accuracy. Furthermore, our proposed 3F2N SNE achieves better overall performance than all other geometry-based SNEs. 

\section*{Supplementary Material}

\subsection{Comparisons on Noisy Synthetic Datasets}
\label{sec.comparison}
To further validate the robustness of 3F2N SNE on noisy range sensor data, we add random Gaussian noise (with respect to three different standard deviations) to our created synthetic depth images. 3F2N SNE is then compared with the state-of-the-art geometry-based SNEs, as shown in Tab. \ref{table.noise_level}. It can be observed that our proposed FD-Median SNE achieves the best results on noisy easy datasets and it performs slightly worse than AngleWeighted SNE \cite{klasing2009comparison} on noisy medium and hard datasets. Referring to Tab. {\color{red}III} in the full paper, AngleWeighted SNE \cite{klasing2009comparison} also performs similarly to FD-Median SNE on clean depth images, but it is extremely computationally intensive. We believe that the noise in depth images affects all SNEs to a similar extent and our proposed method is still among the best geometry-based SNEs.

\begin{figure}[!t]
	\centering
	\includegraphics[width=0.47\textwidth]{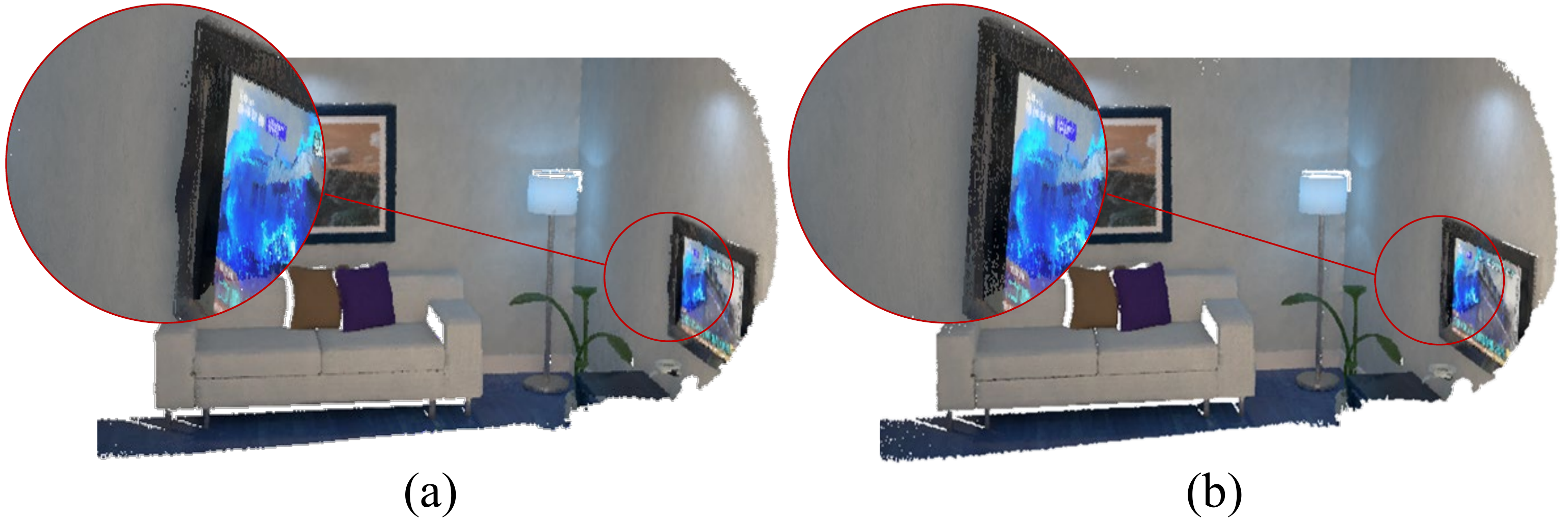}
	\caption{3D scene reconstruction comparison: (a) conventional 3D scene reconstruction; (b) 3D scene reconstruction aided by 3F2N SNE. }
	\label{fig.slam}
\end{figure}
\subsection{Applications of 3F2N SNE}
\label{sec.applications}

As mentioned in the full paper, 3F2N SNE can be applied in a variety of computer vision and robotics tasks. Therefore, in our experiments, we first utilize an off-the-shelf registration algorithm provided by the point cloud
library\footnote{\url{pointclouds.org}} (PCL) to match the 3D point cloud generated from each depth image with a global 3D geometry model. The sensor poses and motion trajectory can then be obtained. Meanwhile, we integrate the surface normal information into the point cloud registration process and acquire another collection of sensor poses and motion trajectory.
Then, we utilize ElasticFusion \cite{whelan2015elasticfusion}, a real-time dense visual SLAM system,  to reconstruct the 3D scenery using the input RGB-D data and two collections of sensor poses and motion trajectories. Two reconstructed 3D scenes are illustrated in Fig. \ref{fig.slam}, where it is obvious that the proposed SNE can improve the 3D geometry reconstruction accuracy.  The experimental results suggest that the 3D reconstruction accuracy can be improved by approximately 19\%, when using the surface normal information  obtained by 3F2N SNE.

Additionally, 3F2N SNE can also be employed to detect planar surfaces through end-to-end semantic segmentation CNNs. As demonstrated in \cite{fan2020sne} and \cite{wang2021tcyb}, fusing surface normal maps and RGB images can bring significant improvements on free-space detection. The high efficiency of 3F2N SNE enables it to be easily deployed in CNNs for surface normal information acquisition, without affecting their training/inference speed.

\balance
{\small
\bibliographystyle{IEEEtran}
\bibliography{egbib}
}


\end{document}